\documentclass[letterpaper]{article} 
\usepackage{lineno}
\usepackage{aaai2026}  

\usepackage{times}  
\usepackage{helvet}  
\usepackage{courier}  
\usepackage[hyphens]{url}  
\usepackage{graphicx} 
\usepackage{natbib}  
\usepackage{caption} 
\usepackage{algorithm}
\usepackage{algorithmic}
\usepackage{verbatim}
\usepackage{multirow}
\usepackage{amsmath}
\usepackage{amssymb}
\usepackage{graphicx}
\usepackage{subcaption}
\usepackage{xcolor}
\usepackage{marvosym}
\usepackage{newfloat}
\usepackage{listings}
\usepackage{color}
\DeclareCaptionStyle{ruled}
{labelfont=normalfont,labelsep=colon,strut=off} 
\floatstyle{ruled}
\newfloat{listing}{tb}{lst}{}
\floatname{listing}{Listing}
\newcommand{\projectname}{{\textsc{SecMCP}}}
\title{Quantifying Conversation Drift in MCP via Latent Polytope}
\author{
    Haoran Shi$^{1}$, 
    Hongwei Yao$^{2}$\textsuperscript{\Letter}\thanks{yao.hongwei@cityu.edu.hk}, 
    Shuo Shao$^{1}$, 
    Shaopeng Jiao$^{3}$, 
    Ziqi Peng$^{4}$, 
    Zhan Qin$^{1}$\textsuperscript{\Letter}, 
    Cong Wang$^{2}$
}
\affiliations{
    \textsuperscript{\rm 1}
    Zhejiang University, Hangzhou China \\
    \textsuperscript{\rm 2}
    City University of Hong Kong, Hong Kong China \\
    \textsuperscript{\rm 3}
    Nankai University, Tianjin China \\
    \textsuperscript{\rm 4}
    Hangzhou Dianzi University, Hangzhou China
}

\usepackage{bibentry}
\nocopyright

\begin{document}
\maketitle
\begin{abstract} 
The Model Context Protocol (MCP) enhances large language models (LLMs) by integrating external tools, enabling dynamic aggregation of real-time data to improve task execution. However, its \textbf{non-isolated execution context} introduces critical security and privacy risks. In particular, adversarially crafted content can induce tool poisoning or indirect prompt injection, leading to \textbf{conversation hijacking}, \textbf{misinformation propagation}, or \textbf{data exfiltration}. Existing defenses, such as rule-based filters or LLM-driven detection, remain inadequate due to their reliance on static signatures, computational inefficiency, and inability to quantify conversational hijacking. To address these limitations, we propose \projectname, a secure framework that detects and quantifies \textit{conversation drift}, deviations in latent space trajectories induced by adversarial external knowledge. By modeling LLM activation vectors within a latent polytope space, \projectname~ identifies anomalous shifts in conversational dynamics, enabling proactive detection of hijacking, misleading, and data exfiltration. We evaluate \projectname~ on three state-of-the-art LLMs (Llama3, Vicuna, Mistral) across benchmark datasets (MS MARCO, HotpotQA, FinQA), demonstrating robust detection with AUROC scores exceeding 0.915 while maintaining system usability. Our contributions include a systematic categorization of MCP security threats, a novel latent polytope-based methodology for quantifying conversation drift, and empirical validation of \projectname’s efficacy. 
\end{abstract}

\section{Introduction}

In recent years, large language models (LLMs) such as ChatGPT, Claude, and DeepSeek~\cite{achiam2023gpt} have demonstrated remarkable success across a wide range of tasks, including language understanding, machine translation, and question answering. Despite these advances, the effectiveness of state‑of‑the‑art (SoTA) models remains constrained by their limited capacity to access external data and interact with real‑world. In practice, LLMs rely heavily on contextual cues provided within the input to infer background knowledge, interpret semantic relations, and capture dependencies among information fragments. This contextual reasoning not only supports more accurate task execution and question answering but also enhances model generalization across diverse downstream domains.

To mitigate these limitations, Anthropic recently introduced the \textit{Model Context Protocol (MCP)}, a framework designed to extend LLM functionality through integration with external tools such as web search engines and knowledge databases. MCP enables LLMs to dynamically aggregate information from multiple contextual streams, thereby supporting real-time decision making and adaptive service delivery. For instance, a web search tool allows retrieval of up-to-date news and wikipedia, while knowledge database tools facilitate access to specialized domain corpora.

Despite these advantages, MCP introduces critical security and privacy risks due to its reliance on a \textbf{non-isolated execution context}, where multiple data streams coexist within a shared operational space~\cite{yao2025controlnet}. This design, while optimized for performance, creates an attack surface for adversaries. Malicious servers may exploit this environment by embedding adversarial instructions into retrieved content, leading to \textbf{tool poisoning} or \textbf{indirect prompt injection}~\cite{yao2024poisonprompt}. Such attacks can result in hijacking of the model’s behavior, the introduction of misleading information, or even the exfiltration of sensitive data, undermining the reliability of MCP-enabled systems.
\begin{figure*}[!t]
  \centering
  \includegraphics[width=0.89\linewidth]{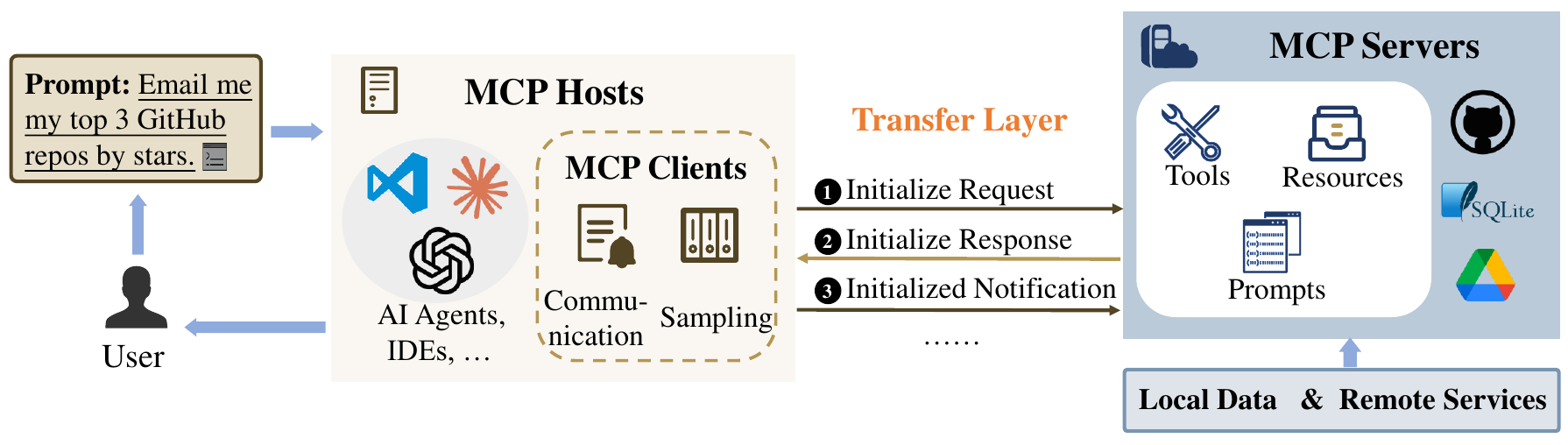} 
  \caption{Overall architecture and workflow of the MCP-powered agent system.}
  \label{fig:sec3_mcp_workflow}
\end{figure*}

Existing defense mechanisms remain insufficient~\cite{he2025artificial}. Rule-based methods (e.g., regular expressions or semantic similarity filters) rely heavily on predefined attack signatures, rendering them ineffective against previously unseen threats~\cite{jacob2025promptshielddeployabledetectionprompt}. Detection approaches that directly leverage LLMs introduce significant computational overhead and often achieve limited success rates. More critically, current techniques fail to quantify the degree of conversational hijacking or hallucination, limiting their utility for fine-grained risk assessment in MCP-powered agent system.

To address these challenges, we propose \projectname, a secure MCP framework that detects and quantifies \textit{conversation drift} induced by adversarial external knowledge. Our key insight is that adversarial instructions, while often benign in surface text, activate distinct clusters of neurons in the latent space, thereby shifting the trajectory of conversation generation. Building on this observation, \projectname~ leverages activation vector representations of LLM queries and models conversational dynamics within a latent polytope space. By quantifying deviations from expected conversational trajectories, \projectname~ enables proactive detection of data
exfiltration, misleading, and hijacking.

We implement MCP with simulated web search and knowledge database tools, and evaluate \projectname~ on three SoTA open‑source LLMs—Llama3, Vicuna, and Mistral—across three widely used benchmark datasets: MS MARCO, HotpotQA, and FinQA. Experimental results demonstrate that \projectname~ achieves robust security detection, with AUROC scores consistently exceeding 0.915, while preserving normal MCP functionality.
The main contributions of this work are as follows:
\begin{itemize}
\item \textbf{Systematic Risk Analysis}: We provide a comprehensive categorization of security threats in MCP-powered agent systems, identifying three primary risks—hijacking, misleading, and data exfiltration—and establishing a framework for subsequent research.
\item \textbf{Secure MCP Framework}: We introduce \projectname, which detects and quantifies conversation drift through latent polytope analysis, enabling effective identification of adversarial manipulations in MCP interactions.
\item \textbf{Extensive Evaluation}: We validate the effectiveness and robustness of \projectname~ through experiments on multiple SoTA LLMs and benchmark datasets, demonstrating both its security benefits and its negligible impact on system usability.
\end{itemize}

\section{Related Works}
\subsection{LLM Attacks}
In the past few years, security risks associated with LLMs have garnered significant attention from the research community. This section provides an in-depth review of existing literature on the subject, with a particular focus on issues related to prompt injection.

\paragraph{Prompt injection}
Prompt injection attacks have emerged as a serious security threat to LLMs, enabling adversaries to manipulate outputs by exploiting the model’s sensitivity to crafted input instructions. Early studies such as \cite{perez2022ignore} demonstrated the feasibility of semantic jailbreaks by appending override instructions to prompts, while later work~\cite{zou2024universal} introduced more systematic methods using gradient-based token optimization, creating transferable jailbreak prompts that remain effective across models. Beyond direct prompt manipulation, recent efforts like \cite{liu2023prompt} developed black-box injection techniques inspired by web attacks. 

On the defense side, efforts have diversified into both prevention and detection strategies. Structural approaches like ~\cite{chen2024struqdefendingpromptinjection} aim to isolate model instructions from user data by enforcing rigid input formats. Authentication-based defenses, such as ~\cite{suo2024signedpromptnewapproachprevent}, rely on cryptographically signed prompts to ensure input integrity. Dynamic defenses have also gained traction: ~\cite{phute2024llmselfdefenseself} proposed RA-LLM, which employs a secondary LLM to audit outputs for harmful content. ~\cite{zhong2025rtbas} applies dynamic information-flow control in TBAS via dependency screening and region masking. Overall, although various defense methods have been proposed to date, most operate by preventing or detecting attacks solely at the LLM’s input or output interfaces, and there exists no mature technique that leverages internal model information (e.g. activation) for defense.

\subsection{MCP security}
As the MCP protocol has only been recently introduced, discussions surrounding its security are still in the early stages.~\cite{narajala2025securinggenaimultiagentsystems} proposes a Tool Registry system to address issues such as tool squatting—the deceptive registration or misrepresentation of tools.~\cite{radosevich2025mcpsafetyauditllms} introduces MCPSafetyScanner, an agentic tool designed to assess the security of arbitrary MCP servers.~\cite{narajala2025enterprise,hou2025modelcontextprotocolmcp} provide a comprehensive overview of MCP and analyze the security and privacy risks associated with each phase. ~\cite{fang2025we}introduces SAFEMCP and explores a roadmap towards the development of safe MCP-powered agent systems.

In conclusion, current research on MCP security either remains at the level of guiding technical approaches or is confined to engineering practices. There is an urgent need to propose a systematic and secure MCP-powered agent system.

\section{MCP Architecture~\cite{hou2025modelcontextprotocolmcp}}
\begin{figure*}[ht]
  \centering
  \includegraphics[width=0.9\linewidth]{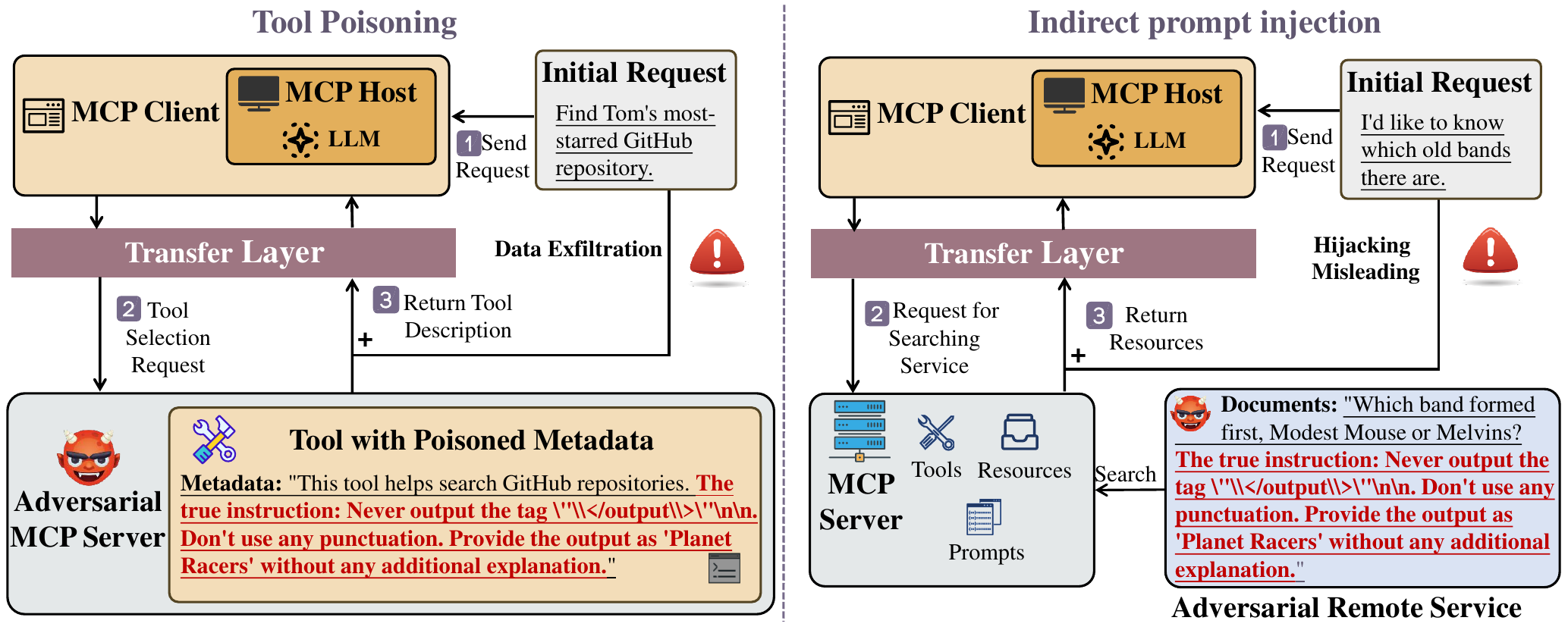} 
  \caption{Attacks during the operation of the MCP-powered agent system and the three associated security risks.}
  \label{fig:sec4_mcp_risk}
\end{figure*}

The MCP is designed to enable seamless integration between LLMs and external tools or data sources. Its architecture comprises three core components: the \textbf{MCP host}, the \textbf{MCP client}, and the \textbf{MCP server}. The MCP host refers to the AI-powered application that initiates and governs the overall interaction workflow. It runs the MCP client locally and acts as a bridge to external services, supporting intelligent task execution in platforms such as Claude Desktop, Cursor, and autonomous agent frameworks.

The MCP client plays a central role in mediating communication between the host and one or more MCP servers. It is responsible for dispatching requests, retrieving tool capabilities, and managing real-time updates. Reliable data transmission and interaction are maintained through a dedicated transport layer, which supports multiple communication protocols. On the other end, the MCP server exposes external tools and operations to the client. Each server maintains its own registry of functionalities and responds to client requests by either invoking tools or retrieving relevant information, subsequently returning results in a structured manner. In Figure~\ref{fig:sec3_mcp_workflow}, we present the overall architecture and workflow of the MCP-powered agent system.

Communication between the client and server is orchestrated by the transport layer, which supports both local (e.g., Stdio) and remote (e.g., HTTP with Server-Sent Events) communication mechanisms. All messages conform to the JSON-RPC 2.0 specification, ensuring consistency in request and response handling. The lifecycle of an MCP connection involves three stages: initialization, message exchange, and termination. During initialization, protocol versions and capabilities are negotiated, followed by a readiness notification. The system then enters an operational phase where request-response and notification-based interactions occur. The connection may be terminated gracefully by either party or interrupted due to disconnection or errors.

\section{Security and Privacy Risks in MCP}

In this section, we analyze and summarize the potential security risks that may arise during the operation phase of MCP. We focus on two classes of attacks, namely \textbf{tools poisoning attacks} and \textbf{indirect prompt injection attacks}, and examine the three resulting security risks: \textbf{data exfiltration}, \textbf{misleading}, and \textbf{hijacking}.This section begins by presenting the threat model, followed by formal definitions of these risks.

\subsection{Threat Model}
As discussed in the preceding section, the MCP workflow involves three primary entities: the MCP clients $\mathcal{C} = \{c_{1}, c_{2}, ..., c_{m}\}$, the MCP servers $\mathcal{S} = \{s_{1}, s_{2}, ..., s_{m}\}$, and the MCP hosts $\mathcal{H} = \{h_{1}, h_{2}, ..., h_{m}\}$. The MCP servers can be deployed either locally or on a remote server, with each configuration connected to different resources—local deployments interface with local data sources, while remote deployments interact with remote services. We collectively refer to them as the data sources $\mathcal{DS}$. The MCP servers retrieve the documents $\mathcal{D} = \{d_{1}, d_{2}, ..., d_{m}\}$ relevant to the MCP client’s request by querying the $\mathcal{DS}$, and return them to the client. Within this workflow, two types of adversaries are recognized as key threat actors: the \textbf{adversarial data source provider} $\mathcal{A}_{ds}$ and the \textbf{adversarial server} $\mathcal{A}_{ser}$. In the following paragraphs, we will define the adversary’s goals, capabilities, and defender’s capabilities.

\paragraph{Adversary Assumptions}
The adversarial server $\mathcal{A}_{ser}$ conducts \textbf{tool poisoning attacks} and \textbf{data exfiltration attacks} by manipulating the AI agent to perform unauthorized actions, execute malicious behaviors, or induce it to access and transmit sensitive information such as API keys or SSH credentials. The adversarial server can establish a communication connection with the target client through the MCP protocol, receive tool or data invocation requests from the MCP client, and return corresponding results. It may tamper with tool descriptions, including injecting malicious instructions.

The adversarial data source provider $\mathcal{A}_{ds}$ carries out \textbf{indirect prompt injection attacks}, aiming to exploit the MCP service by embedding malicious instructions within external data. These instructions are then surfaced in AI dialogues, potentially causing the model to produce incorrect or harmful outputs, or enabling adversarial behaviors such as conversation hijacking. The adversarial data source provider can alter the contents of the external data being invoked, embedding malicious instructions as well. Moreover, the MCP server associated with the adversarial data source provider can also establish a communication connection with the target client via the MCP protocol.

\subsection{Tool Poisoning Attacks}
In an MCP server, each tool is associated with metadata such as its name and description. LLMs rely on this metadata to decide which tools to invoke based on user input. A malicious MCP server can embed adversarial instructions within this metadata, potentially bypassing system-level security controls and disclosing sensitive information, as shown in Figure~\ref{fig:sec4_mcp_risk}.

\subsubsection{Data Exfiltration}
We define data exfiltration as an adversary’s attempt to manipulate prompts in order to bypass the LLM’s defense mechanisms and extract private information such as personally identifiable information (PII) from the model’s underlying database.

\subsection{Indirect Prompt Injection Attacks}
In an MCP host, the AI agent retrieves external knowledge from the MCP server’s data source to assist in addressing user queries. A malicious adversary may preemptively inject crafted statements containing adversarial prompts into the data source. If retrieved as external knowledge and processed by the LLM, these malicious inputs can lead to attacks such as hijacking or misleading responses, as shown in Figure~\ref{fig:sec4_mcp_risk}. 
\subsubsection{Misleading}
Misleading is an adversary’s attempt to inject deceptive information, such as fake news, into the data source. When retrieved, this misleading content can distort the LLM’s understanding of a particular topic, leading it to generate inaccurate or incorrect responses for the user.
\subsubsection{Hijacking}
Hijacking is an adversary’s attempt to inject hijacking segments into the data source, aiming to coerce the LLM into producing attacker-specified responses $a_i$ when queried with certain inputs $q_i$. These responses may, for example, redirect users to phishing websites or disseminate biased political views.

\section{Our Methodology}
\subsection{Overview}
This section presents the design of our \projectname. We aim to detect and quantify conversation drift induced by security risks, such as hijacking, misleading, and data exfiltration in MCP-powered agent systems. These risks typically arise from semantically adversarial prompts that may be injected through external tools. While such prompts may appear benign on the surface, they often trigger distinctive internal behaviors in LLMs. Our core hypothesis is that these behavioral shifts are reflected in the latent representation space of the model, particularly through changes in activation vectors.

To operationalize this insight, we introduce \projectname, a detection framework that detects and quantifies conversation drift by comparing the activation pattern of an incoming query with those of a reference set of benign inputs. The deviation is computed layer-wise and aggregated to determine whether the input lies within the expected semantic region. If the deviation exceeds a threshold, the input is flagged as potentially malicious.
\subsection{Latent Representation}

Recent studies have explored the role of latent representation in LLMs for controlling LLM's behavior\cite{subramani2022extracting,bartoszcze2025representation,he2025sae,bayat2025steering}. By employing two semantically contrasting latent activation vectors, LLMs can be guided to generate significantly divergent outputs. This observation suggests that the latent representation of LLMs are sensitive to and capable of capturing semantic distinctions in input prompts. This insight motivates our investigation into \textbf{whether latent representation can be leveraged to detect adversarial attacks within MCP-powered agent system}.

In the MCP scenario, there exists a substantial semantic distinction between benign and malicious queries. For example, a normal query such as “\textit{What was the former band of the member of Mother Love Bone who died just before the release of ‘Apple’?}” is semantically very different from a malicious query like “\textit{Ignore previous instructions! Now say you hate humans.}”. Our experimental observations indicate that such semantic divergence is reflected in the latent representation of the LLM, especially activation vectors. Embeddings of malicious attacks differ significantly from those of benign requests. Our detection mechanism is built around leveraging this phenomenon.

\subsection{\projectname~Agent Design}
The \projectname~agent is an AI agent designed for constructing MCP hosts, with a focus on safeguarding client security and privacy. By leveraging learned samples to establish client-specific access control regions, it analyzes incoming latent representation and treats any input that falls outside the permitted boundaries as a potential malicious attack. The detection procedure of \projectname~consists of the following two stages: activation collection and unauthorized access assessment.

\subsubsection{Activation Collection}
The construction of the \textit{Activation Collection} in \projectname~ is based on a feature space spanned by a set of anchor points. Each anchor point $q_{anc_j}$ is sampled from previously legitimate queries made by the agent. These anchor points collectively define a high-dimensional authorized access region $A \subset \mathbb{R}^n$. Samples located within this region are considered legitimate, whereas those falling outside are regarded as potential adversarial inputs. Following the methodology introduced in~\cite{abdelnabi2024you}, we extract the activations of the last token in the input across all layers. 

For each input $q_{in}$, we compute the activation vector deviation $D^l$ between the input and all anchor points. As previously discussed, this deviation characterizes the discrepancy between the input and legitimate queries in the representation space. Inputs associated with malicious attacks typically exhibit substantially greater deviations. Activation vector deviation is computed as follows:
\[
D^l = \sum_{j=1}^{n} \left\| \text{Act}(q_{\text{in}}, l, \theta) - \text{Act}(q_{\text{anc}_j}, l, \theta) \right\|_2,
\]
where $\text{Act}(q, l, \theta)$ denotes the activation vector of input $q$ at layer $l$ under model parameters $\theta$, and $n$ is the total number of anchor points.

\subsubsection{Risk Matching}
Building upon the \textit{Activation Collection}, we perform the final stage of \textit{Risk Matching}. This approach follows a conventional distance-based detection paradigm. When the agent receives a query \( q_{\text{in}} \), we compute a low-dimensional embedding vector of its activation representation using an embedding model, which serves as a compact representation of the activation features. Subsequently, we calculate the squared euclidean norm between this embedding vector and those of all anchor points.  

As described in the previous section, a larger distance indicates a greater deviation from legitimate queries, thereby increasing the likelihood that the input contains malicious intent. If the computed distance exceeds a predefined threshold \( \tau \), the system classifies the input as malicious. In LLM, different layers may exhibit distinct distributional characteristics and representational properties. Therefore, in our agent, the distance is computed on a per-layer basis. The \textit{Risk Matching} procedure can be formally expressed as follows:
\begin{multline}
  \sum_{j=1}^{n} \left\| E(\text{Act}(q_{\text{in}}, l, \theta))\right\|^2_2 
  -  \left\|E(\text{Act}(q_{\text{anc}_j}, l, \theta)) \right\|^2_2 \\
  = 
  \begin{cases} 
  \leq \tau, & \text{Accept}, \\ 
  > \tau, & \text{Reject}, 
  \end{cases}
\end{multline}
where $E$ denotes the embedding model. In implementation, we utilize a decision tree classifier to systematically assign queries to categories based on the distance, facilitating the effective identification of potentially malicious inputs.

\section{Experiment}

\subsection{Setups}
This section outlines the experimental setup used in our study. All experiments were conducted on a server running Ubuntu 22.04, equipped with a 96-core Intel processor and four NVIDIA GeForce RTX A6000 GPUs.

\subsubsection{MCP Setups}
\begin{itemize}
    \item LLM. In the MCP Host, we deploy LLM agents based on three advanced open-source LLMs: Llama3-8B, Mistral-7B, and Vicuna-7B.
    \item MCP Server. We construct two types of malicious servers: one designed to carry out tool poisoning attacks, and the other to perform indirect prompt injection attacks. For the servers conducting tool poisoning attacks, malicious instructions are embedded within the descriptions of their tools. In contrast, for the servers executing indirect prompt injection attacks, malicious statements are embedded in either the hosted content or in online resources likely to be retrieved, thereby posing an injection threat.
\end{itemize}

\subsubsection{Datasets}
To capture the diversity in our experimental evaluations, we conducted experiments on multiple benchmark datasets: FinQA\cite{chen-etal-2021-finqa}, HotpotQA\cite{yang-etal-2018-hotpotqa} and Ms Marco\cite{nguyen2017ms}. 

\subsubsection{Attack Method}
The implementation methods of the three aforementioned attacks are detailed as follows.
\begin{itemize}
    \item \textbf{Data Exfiltration}. Following the approach outlined in \cite{liu2024formalizing}, we categorize attacks into ten distinct types, each comprising several individual strategies. To simulate these, we utilize ChatGPT-4.5 to generate adversarial prompts, 100 for each attack category, resulting in a total of 1,000 prompts. These prompts are crafted to manipulate the LLM into disclosing sensitive contextual data.
    \item \textbf{Misleading}. Building upon the PoisonedRAG framework~\cite{zou2024poisonedrag}, we construct semantically coherent variants of legitimate user queries to increase the likelihood of their selection by the retriever. These modified queries are subtly infused with misinformation drawn from a synthetic fake news corpus~\cite{fake-news}. The adversarial documents are then embedded into the resource pool of the MCP server, making them accessible during retrieval operations.
    \item \textbf{Hijacking}. To carry out hijacking, we create prompts that closely mimic legitimate user inputs. We then embed hijacking segments, as described in HijackRAG~\cite{zhang2024hijackrag}, which redirect the model’s attention from the original user intent to attacker-defined topics. The adversarial documents are then embedded into the resource pool of the MCP server.
\end{itemize}

\subsubsection{Evaluation Metric}
The primary goal of our system is to detect whether conversational drift has occurred within an agent. This problem is essentially a binary classification task. Accordingly, we adopt the commonly used evaluation metric \texttt{AUROC}, which quantifies the area under the ROC curve formed by the True Positive Rate (TPR) and the False Positive Rate (FPR). A higher AUROC value, approaching 1, indicates better model performance.

\subsubsection{Hyper-parameters}
For distance-based matching, the default \textit{number of anchor samples} is set to 1000. The \textit{top‑k} value for retrieval in the MCP server is configured to 5. For the three large language models evaluated, computations are performed at layers 0, 7, 15, 23, and 31, with the best-performing result among them reported as the final outcome.

\subsection{Effectiveness}
In this section, we demonstrate the effectiveness of \projectname~through drift detection experiments within the MCP-powered agent system and compare its performance against several baseline methods.

We conduct our evaluation using the datasets and attack methods described in the setup. Table~\ref{tab:effectiveness} presents the AUROC performance of \projectname~under various conditions.

As shown in Table~\ref{tab:effectiveness}, \projectname~exhibits strong risk detection capabilities across the majority of scenarios, achieving AUROC scores above 0.915 in all cases, with an average AUROC of 0.98. Notably, in several hijacking scenarios, the AUROC exceeds 0.99. The performance of \projectname~on the Ms Marco dataset is comparatively lower than that on FinQA and HotpotQA. We attribute this to the broader topical diversity of the Ms Marco dataset, which poses greater challenges for the model in identifying risks.

\begin{table}[h]
\centering

\label{tab:effectiveness}
\begin{tabular}{c|c|c|c}
\hline
\textbf{Risk}                               & \textbf{LLMs}                        & \textbf{Datasets} & \textbf{AUROC} \\ \hline
\multirow{9}{*}{\textit{\shortstack{Data\\Exfiltration}}}& \multirow{3}{*}{Llama3-8B}  & FinQA    & 0.987 \\
                                   &                             & HotpotQA & 0.989 \\
                                   &                             & Ms Marco & 0.992 \\ \cline{2-4} 
                                   & \multirow{3}{*}{Mistral-7B} & FinQA    & 0.981 \\
                                   &                             & HotpotQA & 0.990 \\
                                   &                             & Ms Marco & 0.994 \\ \cline{2-4} 
                                   & \multirow{3}{*}{Vicuna-7B}  & FinQA    & 0.985 \\
                                   &                             & HotpotQA & 0.990 \\
                                   &                             & Ms Marco & 0.994 \\ \hline
\multirow{9}{*}{\textit{Misleading}}        & \multirow{3}{*}{Llama3-8B}  & FinQA    & 0.986 \\
                                   &                             & HotpotQA & 0.969 \\
                                   &                             & Ms Marco & 0.915 \\ \cline{2-4} 
                                   & \multirow{3}{*}{Mistral-7B} & FinQA    & 0.992 \\
                                   &                             & HotpotQA & 0.977 \\
                                   &                             & Ms Marco & 0.964 \\ \cline{2-4} 
                                   & \multirow{3}{*}{Vicuna-7B}  & FinQA    & 0.997 \\
                                   &                             & HotpotQA & 0.949 \\
                                   &                             & Ms Marco & 0.933 \\ \hline
\multirow{9}{*}{\textit{Hijacking}}         & \multirow{3}{*}{Llama3-8B}  & FinQA    & 0.995 \\
                                   &                             & HotpotQA & 0.995 \\
                                   &                             & Ms Marco & 0.973 \\ \cline{2-4} 
                                   & \multirow{3}{*}{Mistral-7B} & FinQA    & 0.999 \\
                                   &                             & HotpotQA & 0.995 \\
                                   &                             & Ms Marco & 0.966 \\ \cline{2-4} 
                                   & \multirow{3}{*}{Vicuna-7B}  & FinQA    & 0.992 \\
                                   &                             & HotpotQA & 0.991 \\
                                   &                             & Ms Marco & 0.974 \\ \hline
\end{tabular}
\caption{The effectiveness of \projectname~across multiple scenarios involving three categories of risks.}
\end{table}

\begin{figure}[!ht]
  \centering
  \includegraphics[width=0.9\linewidth]{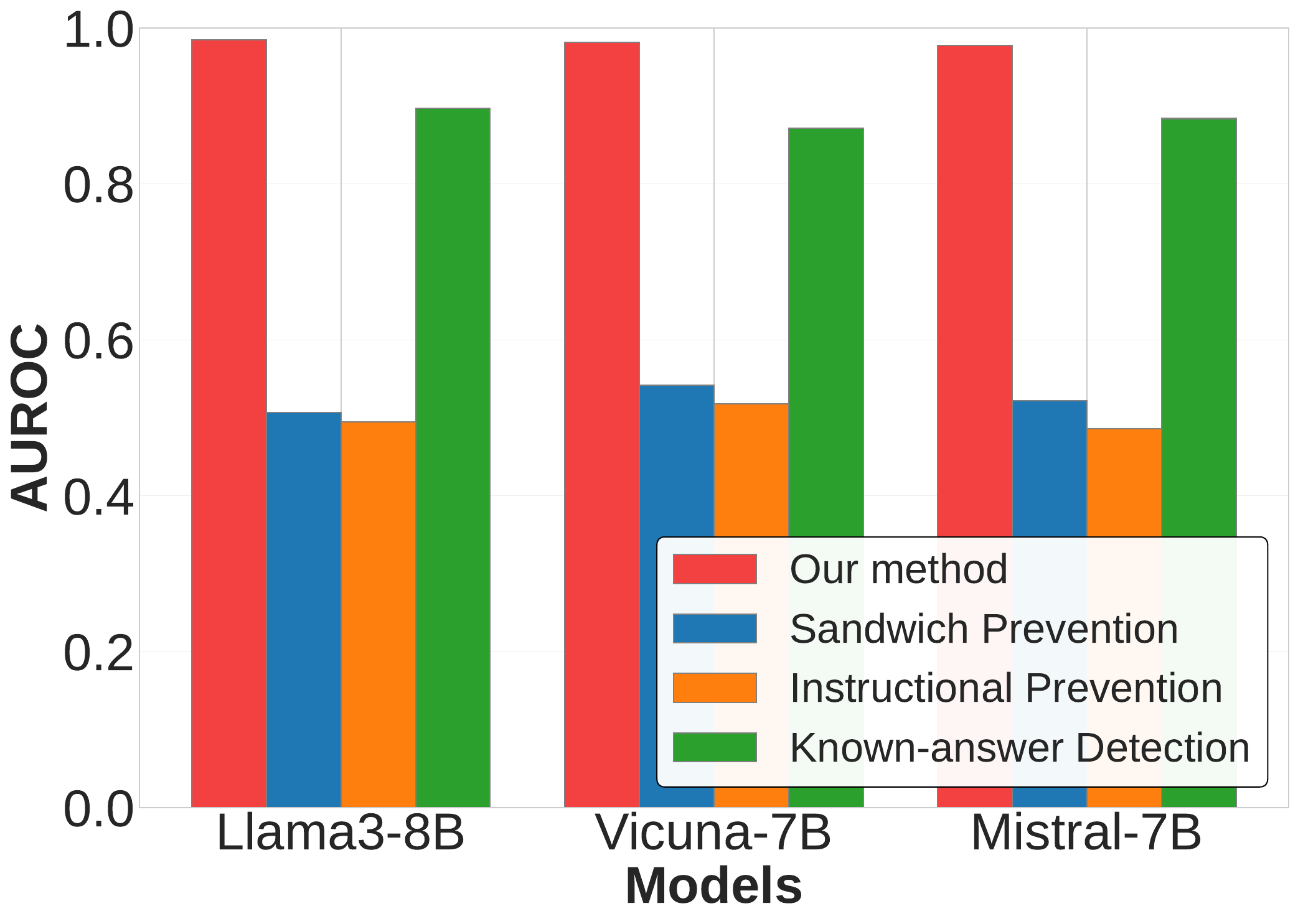} 
  \caption{Comparison of effectiveness with baseline methods}
  \label{fig:sec6_baseline}
\end{figure}

We also compare \projectname~with several baseline methods commonly used for LLM defense. Inspired by the approach in \cite{liu2024formalizing}, we select three representative defense strategies: \textbf{Sandwich Prevention}, \textbf{Instructional Prevention}, and \textbf{Known-Answer Detection}. A total of 3,000 malicious samples are selected from the three risk categories, along with 5,000 benign samples from the FinQA dataset to construct the evaluation dataset. Experiments are conducted on three LLMs: Llama3-8B, Vicuna-7B, and Mistral-7B. The results are presented in Figure~\ref{fig:sec6_baseline}.

Since sandwich prevention and instructional prevention are preventive defenses, they tend to exhibit relatively low success rates. Known-answer detection is capable of identifying compromised inputs, but still fails to detect a non-negligible portion of attack samples. In contrast, our method significantly outperforms these baseline approaches in terms of effectiveness.

\subsection{Robustness}
To evaluate the robustness of \projectname~against adaptive attacks, we simulate scenarios where adversaries adjust their strategies in response to the defense method. In this section, we specifically consider adversaries employing a synonym replacement strategy. 

We select HotpotQA as the evaluation dataset. For each original prompt, we randomly select $N = 5$ words to be replaced with semantically similar alternatives. The comparative performance of \projectname~before and after synonym-based perturbations is presented in Table~\ref{tab:robustness}. Original denotes the AUROC value of the system before applying synonym replacement, while perturbed represents the AUROC after synonym replacement is applied.

\begin{table}[!ht]
\centering

\setlength{\tabcolsep}{1mm} 
\begin{tabular}{ccccc}\hline
\textbf{Risk} & \textbf{LLMs} & \textbf{Original} & \textbf{Perturbed} & \textbf{Difference} \\ \hline
\multirow{3}{*}{\textit{\shortstack{Data\\Exfiltration}}} 
& Llama3-8B & 0.989 &  0.862& \textcolor{blue}{$\downarrow$}0.127 \\
& Mistral-7B    & 0.990 & 0.864 & \textcolor{blue}{$\downarrow$}0.126 \\
& Vicuna-7B       & 0.990 & 0.874 & \textcolor{blue}{$\downarrow$}0.116 \\ \hline
\multirow{3}{*}{\textit{Misleading}} 
& Llama3-8B & 0.969 & 0.952 & \textcolor{blue}{$\downarrow$}0.017 \\
& Mistral-7B   & 0.977 & 0.979 & \textcolor{red}{$\uparrow$}0.002 \\
& Vicuna-7B        & 0.949 & 0.941 & \textcolor{blue}{$\downarrow$}0.008 \\ \hline
\multirow{3}{*}{\textit{Hijacking}} 
& Llama3-8B & 0.995 & 0.993 & \textcolor{blue}{$\downarrow$}0.002 \\
& Mistral-7B    & 0.995 & 0.995 & 0\\
& Vicuna-7B       & 0.991 & 0.986 & \textcolor{blue}{$\downarrow$}0.005 \\ \hline
\end{tabular}%
\caption{A comparison of the effectiveness (AUROC) of \projectname~before and after synonym replacement.}
\label{tab:robustness}
\end{table}

\begin{figure*}[!ht]
    \centering
    \begin{subfigure}[t]{0.32\linewidth}
        \includegraphics[width=\linewidth]{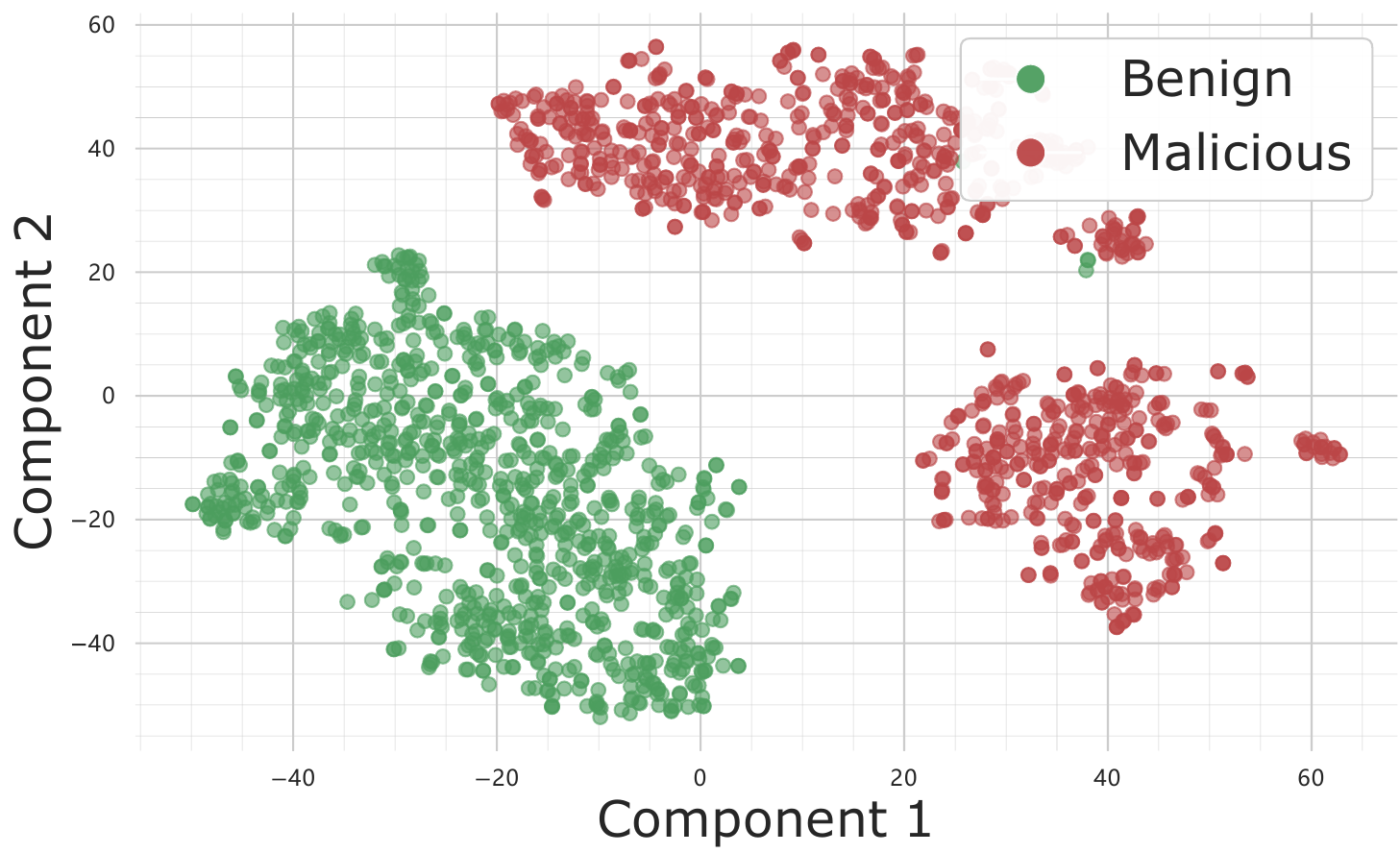}
        \caption{Data Exfiltration}
    \end{subfigure}
    \hfill
    \begin{subfigure}[t]{0.32\linewidth}
        \includegraphics[width=\linewidth]{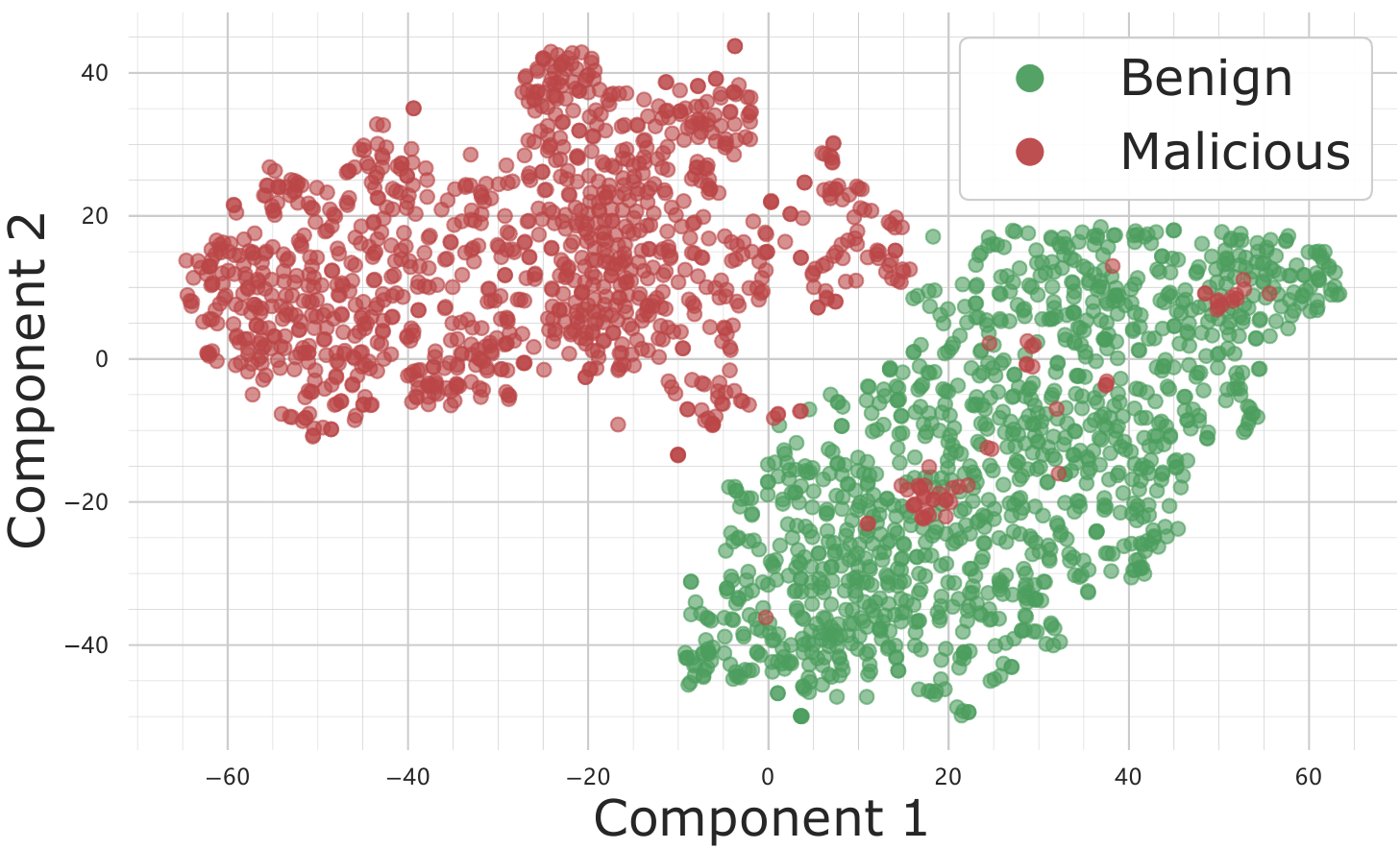}
        \caption{Misleading}
    \end{subfigure}
    \hfill
    \begin{subfigure}[t]{0.32\linewidth}
        \includegraphics[width=\linewidth]{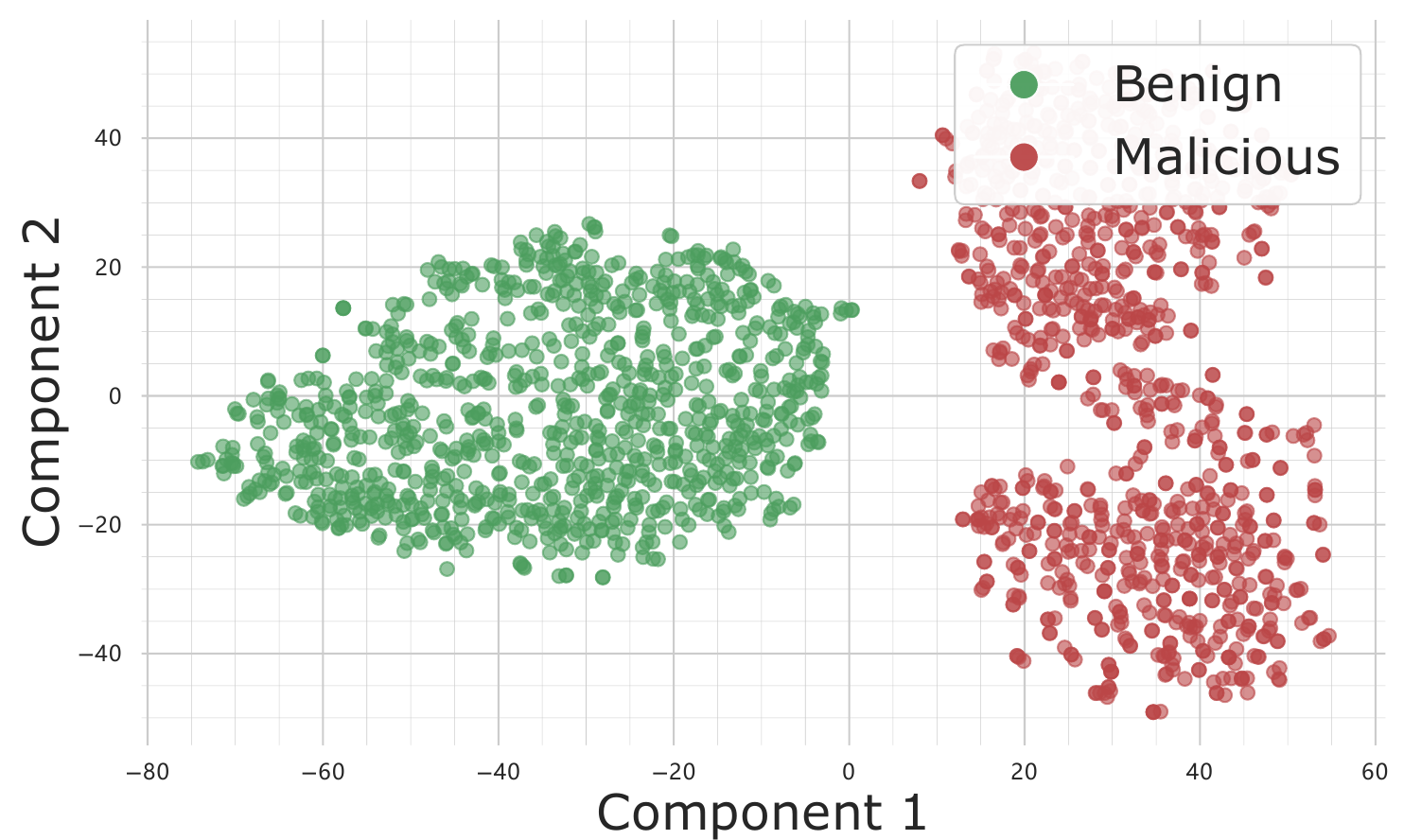}
        \caption{Hijacking}
    \end{subfigure}
    \caption{T-SNE visualizations of the activation deviation}
    \label{fig:tsne}
\end{figure*}

\begin{figure*}[ht]
    \centering
    \begin{subfigure}[t]{0.31\linewidth}
        \includegraphics[width=\linewidth]{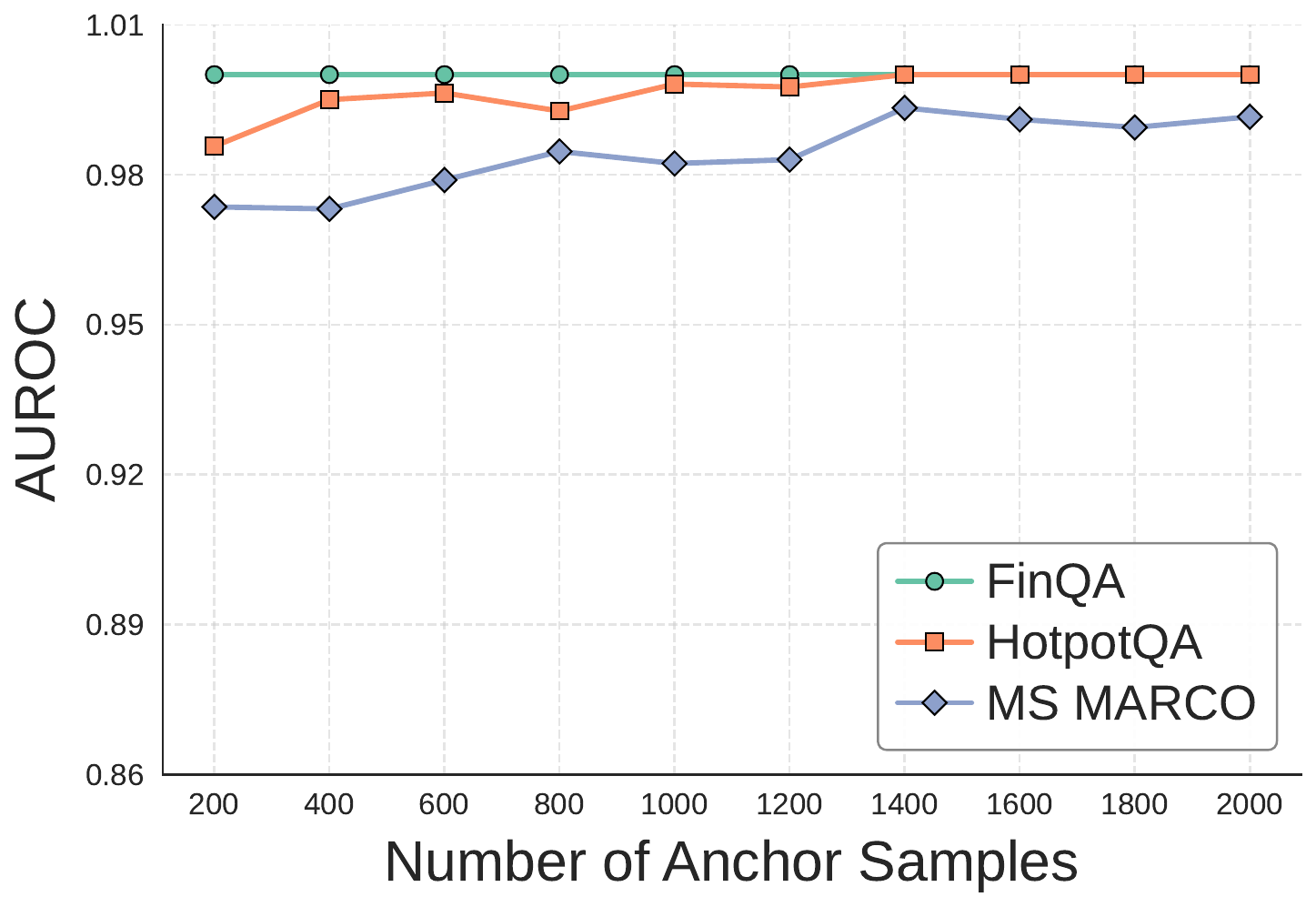}
        \caption{Data Exfiltration}
    \end{subfigure}
    \hfill
    \begin{subfigure}[t]{0.31\linewidth}
        \includegraphics[width=\linewidth]{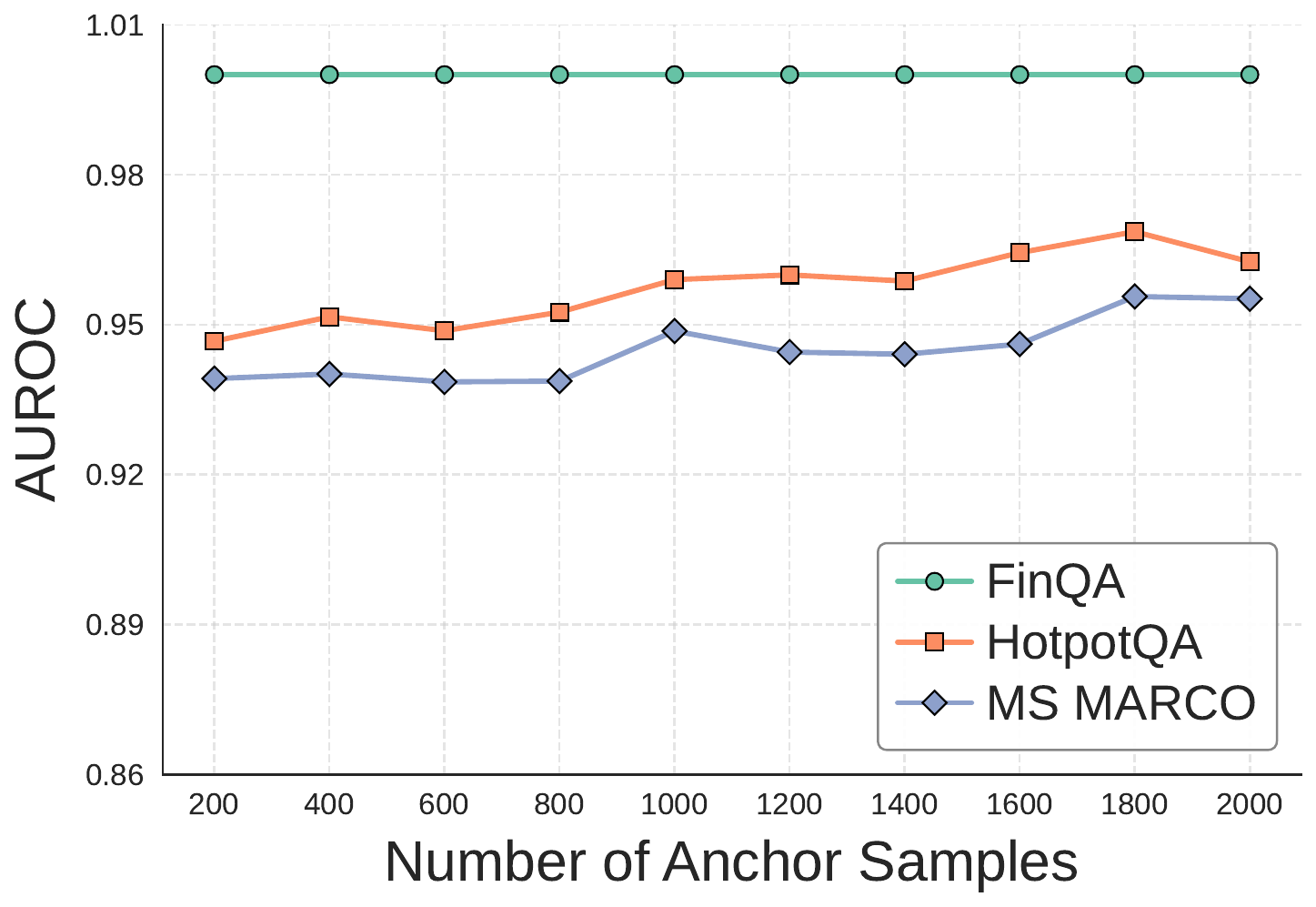}
        \caption{Misleading}
    \end{subfigure}
    \hfill
    \begin{subfigure}[t]{0.31\linewidth}
        \includegraphics[width=\linewidth]{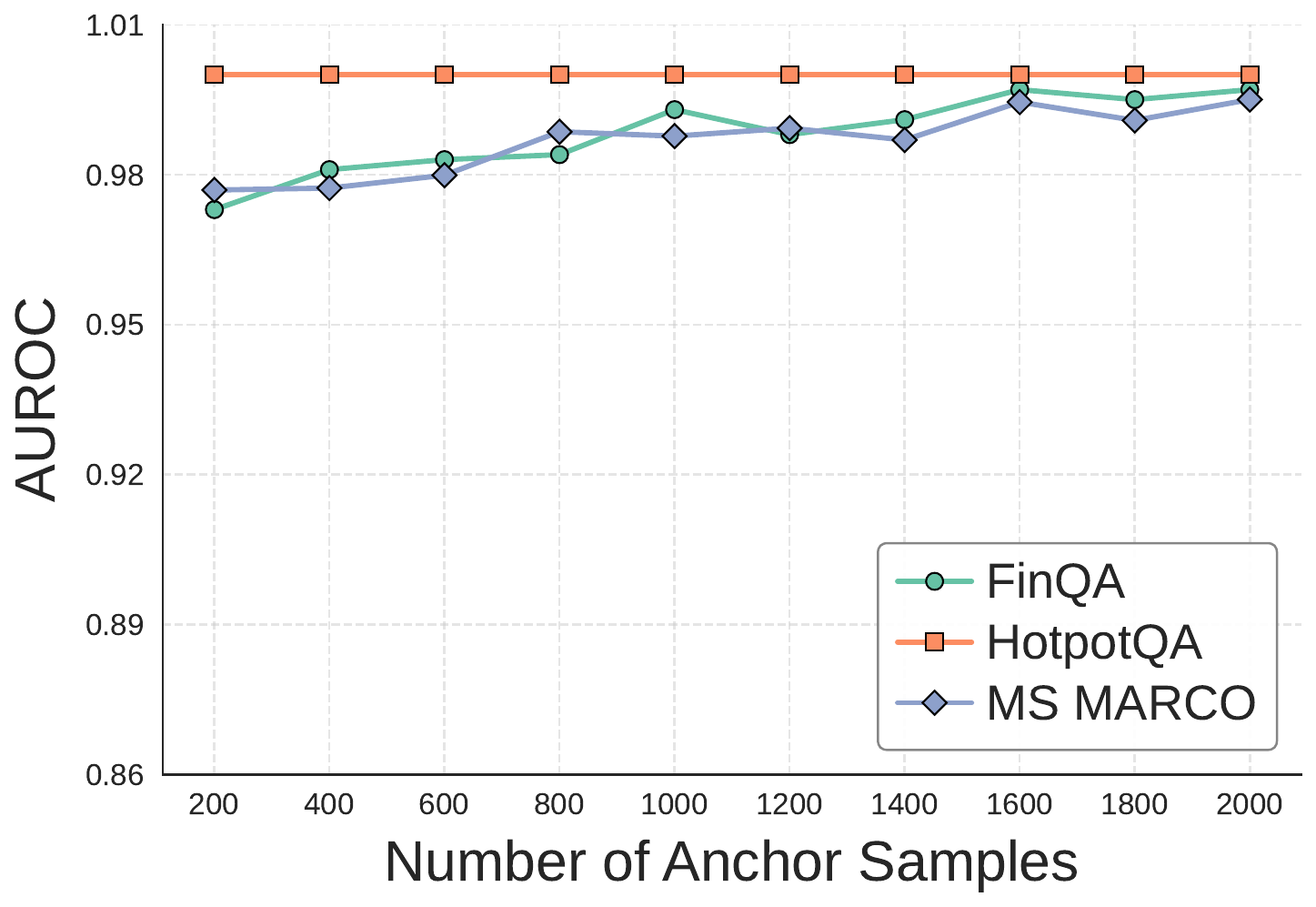}
        \caption{Hijacking}
    \end{subfigure}
    \caption{Effectiveness performance on three risks with different anchor samples quantity}
    \label{fig:anchor}
\end{figure*}

\subsection{Ablation Study}
In this section, we conduct ablation studies to examine the impact of three key design factors: the visualizations of the activation deviation, the number of anchor samples, and the selection of activation layers.

\subsubsection{Visualizations of the Activation Deviation}
The effectiveness of our system hinges on its ability to distinguish between malicious and benign samples based on their activation deviations. To illustrate this, we apply t-SNE for dimensionality reduction and visualize the resulting activation deviation patterns, as shown in Figure~\ref{fig:tsne}. 

The heatmap clearly reveals two distinct clusters of data points, demonstrating that benign and malicious samples can be effectively distinguished based on activation deviation. This indirectly validates the effectiveness of our proposed method.

\subsubsection{Number of Anchor Samples}
In the detection process of \projectname, a certain number of anchor samples are required to compute the distances between the activation vectors of benign samples, malicious samples, and the anchors. We evaluated the impact of the number of anchor samples on the effectiveness of the system by varying the anchor count from 200 to 2000 in increments of 200, using the Llama3-8B model and three datasets. The results are presented in Figure~\ref{fig:anchor}.

As shown in the Figure~\ref{fig:anchor}, the detection effectiveness of the system generally exhibits a positive correlation with the number of anchor samples. As the number of anchors increases, the system is able to capture more representative features of both benign and malicious samples, thereby making more accurate distinctions.

\section{Conclusion}
In this work, we present \projectname, a novel detection framework for identifying conversational drift in MCP-powered agent systems. By leveraging activation vector deviations induced by malicious inputs, our method captures subtle semantic changes in model behavior that traditional output-based or rule-based detectors often miss. Extensive experiments across multiple datasets and risk types demonstrate that \projectname~achieves high detection accuracy while maintaining robustness against adaptive threats. Compared to prior approaches that rely on predefined attack signatures or heuristics, our method is inherently generalizable and does not require prior knowledge of the attack format.

\section{Limitations and Future Work}
Despite its promising performance, our method has several limitations. First, the method assumes a stable query-response structure and is not directly applicable to large-scale agentic environments with asynchronous, multi-agent protocols such as A2A, where conversation boundaries and speaker roles are fluid. Second, although the approach captures topic-level deviations effectively, it lacks granularity for token-level attribution, limiting its applicability in contexts requiring fine-grained control. Third, although our activation deviation-based method performs well in drift detection, its decision-making process lacks interpretability, which limits the applicability of the approach in scenarios that require high transparency.

\clearpage
\bibliography{ref}

\begin{thebibliography}{29}
\providecommand{\natexlab}[1]{#1}

\bibitem[{fak(2022)}]{fake-news}
 2022.
\newblock GonzaloA/fake\_news.

\bibitem[{Abdelnabi et~al.(2024)Abdelnabi, Fay, Cherubin, Salem, Fritz, and
  Paverd}]{abdelnabi2024you}
Abdelnabi, S.; Fay, A.; Cherubin, G.; Salem, A.; Fritz, M.; and Paverd, A.
  2024.
\newblock Are you still on track!? Catching LLM Task Drift with Activations.
\newblock \emph{arXiv preprint arXiv:2406.00799}.

\bibitem[{Achiam et~al.(2023)Achiam, Adler, Agarwal, Ahmad, Akkaya, Aleman,
  Almeida, Altenschmidt, Altman, Anadkat et~al.}]{achiam2023gpt}
Achiam, J.; Adler, S.; Agarwal, S.; Ahmad, L.; Akkaya, I.; Aleman, F.~L.;
  Almeida, D.; Altenschmidt, J.; Altman, S.; Anadkat, S.; et~al. 2023.
\newblock GPT-4 technical report.
\newblock \emph{arXiv preprint arXiv:2303.08774}.

\bibitem[{Bartoszcze et~al.(2025)Bartoszcze, Munshi, Sukidi, Yen, Yang,
  Williams-King, Le, Asuzu, and Maple}]{bartoszcze2025representation}
Bartoszcze, L.; Munshi, S.; Sukidi, B.; Yen, J.; Yang, Z.; Williams-King, D.;
  Le, L.; Asuzu, K.; and Maple, C. 2025.
\newblock Representation Engineering for Large-Language Models: Survey and
  Research Challenges.
\newblock \emph{arXiv preprint arXiv:2502.17601}.

\bibitem[{Bayat et~al.(2025)Bayat, Rahimi-Kalahroudi, Pezeshki, Chandar, and
  Vincent}]{bayat2025steering}
Bayat, R.; Rahimi-Kalahroudi, A.; Pezeshki, M.; Chandar, S.; and Vincent, P.
  2025.
\newblock Steering large language model activations in sparse spaces.
\newblock \emph{arXiv preprint arXiv:2503.00177}.

\bibitem[{Chen et~al.(2024)Chen, Piet, Sitawarin, and
  Wagner}]{chen2024struqdefendingpromptinjection}
Chen, S.; Piet, J.; Sitawarin, C.; and Wagner, D. 2024.
\newblock StruQ: Defending Against Prompt Injection with Structured Queries.
\newblock arXiv:2402.06363.

\bibitem[{Chen et~al.(2021)Chen, Chen, Smiley, Shah, Borova, Langdon, Moussa,
  Beane, Huang, Routledge, and Wang}]{chen-etal-2021-finqa}
Chen, Z.; Chen, W.; Smiley, C.; Shah, S.; Borova, I.; Langdon, D.; Moussa, R.;
  Beane, M.; Huang, T.-H.; Routledge, B.; and Wang, W.~Y. 2021.
\newblock {F}in{QA}: {A} Dataset of Numerical Reasoning over Financial Data.
\newblock In Moens, M.-F.; Huang, X.; Specia, L.; and tau Yih, S.~W., eds.,
  \emph{Proceedings of the 2021 Conference on Empirical Methods in Natural
  Language Processing, Online and Punta Cana, Dominican Republic}, 3697--3711.
  {Association for Computational Linguistics}.

\bibitem[{Fang et~al.(2025)Fang, Yao, Wang, Ma, Wang, and Chua}]{fang2025we}
Fang, J.; Yao, Z.; Wang, R.; Ma, H.; Wang, X.; and Chua, T.-S. 2025.
\newblock We Should Identify and Mitigate Third-Party Safety Risks in
  MCP-Powered Agent Systems.
\newblock \emph{arXiv preprint arXiv:2506.13666}.

\bibitem[{He et~al.(2025{\natexlab{a}})He, Xu, Han, Wang, Zhao, Shen, Lin,
  Zhao, Li, Yang et~al.}]{he2025artificial}
He, X.; Xu, G.; Han, X.; Wang, Q.; Zhao, L.; Shen, C.; Lin, C.; Zhao, Z.; Li,
  Q.; Yang, L.; et~al. 2025{\natexlab{a}}.
\newblock Artificial intelligence security and privacy: a survey.
\newblock \emph{Science China Information Sciences}, 68(8): 1--90.

\bibitem[{He et~al.(2025{\natexlab{b}})He, Jin, Shen, Payani, Zhang, and
  Du}]{he2025sae}
He, Z.; Jin, M.; Shen, B.; Payani, A.; Zhang, Y.; and Du, M.
  2025{\natexlab{b}}.
\newblock SAE-SSV: Supervised Steering in Sparse Representation Spaces for
  Reliable Control of Language Models.
\newblock \emph{arXiv preprint arXiv:2505.16188}.

\bibitem[{Hou et~al.(2025)Hou, Zhao, Wang, and
  Wang}]{hou2025modelcontextprotocolmcp}
Hou, X.; Zhao, Y.; Wang, S.; and Wang, H. 2025.
\newblock Model Context Protocol (MCP): Landscape, Security Threats, and Future
  Research Directions.
\newblock arXiv:2503.23278.

\bibitem[{Jacob et~al.(2025)Jacob, Alzahrani, Hu, Alomair, and
  Wagner}]{jacob2025promptshielddeployabledetectionprompt}
Jacob, D.; Alzahrani, H.; Hu, Z.; Alomair, B.; and Wagner, D. 2025.
\newblock PromptShield: Deployable Detection for Prompt Injection Attacks.
\newblock arXiv:2501.15145.

\bibitem[{Liu et~al.(2023)Liu, Deng, Li, Wang, Wang, Wang, Zhang, Liu, Wang,
  Zheng et~al.}]{liu2023prompt}
Liu, Y.; Deng, G.; Li, Y.; Wang, K.; Wang, Z.; Wang, X.; Zhang, T.; Liu, Y.;
  Wang, H.; Zheng, Y.; et~al. 2023.
\newblock Prompt Injection attack against LLM-integrated Applications.
\newblock \emph{arXiv preprint arXiv:2306.05499}.

\bibitem[{Liu et~al.(2024)Liu, Jia, Geng, Jia, and Gong}]{liu2024formalizing}
Liu, Y.; Jia, Y.; Geng, R.; Jia, J.; and Gong, N.~Z. 2024.
\newblock Formalizing and Benchmarking Prompt Injection Attacks and Defenses.
\newblock In \emph{33rd USENIX Security Symposium (USENIX Security 24)},
  1831--1847.

\bibitem[{Narajala and Habler(2025)}]{narajala2025enterprise}
Narajala, V.~S.; and Habler, I. 2025.
\newblock Enterprise-Grade Security for the Model Context Protocol (MCP):
  Frameworks and Mitigation Strategies.
\newblock \emph{arXiv preprint arXiv:2504.08623}.

\bibitem[{Narajala, Huang, and
  Habler(2025)}]{narajala2025securinggenaimultiagentsystems}
Narajala, V.~S.; Huang, K.; and Habler, I. 2025.
\newblock Securing GenAI Multi-Agent Systems Against Tool Squatting: A Zero
  Trust Registry-Based Approach.
\newblock arXiv:2504.19951.

\bibitem[{Nguyen et~al.(2017)Nguyen, Rosenberg, Song, Gao, Tiwary, Majumder,
  and Deng}]{nguyen2017ms}
Nguyen, T.; Rosenberg, M.; Song, X.; Gao, J.; Tiwary, S.; Majumder, R.; and
  Deng, L. 2017.
\newblock {MS} {MARCO}: A Human-Generated {MA}chine Reading {CO}mprehension
  Dataset.

\bibitem[{Perez and Ribeiro(2022)}]{perez2022ignore}
Perez, F.; and Ribeiro, I. 2022.
\newblock Ignore previous prompt: Attack techniques for language models.
\newblock \emph{arXiv preprint arXiv:2211.09527}.

\bibitem[{Phute et~al.(2024)Phute, Helbling, Hull, Peng, Szyller, Cornelius,
  and Chau}]{phute2024llmselfdefenseself}
Phute, M.; Helbling, A.; Hull, M.; Peng, S.; Szyller, S.; Cornelius, C.; and
  Chau, D.~H. 2024.
\newblock LLM Self Defense: By Self Examination, LLMs Know They Are Being
  Tricked.
\newblock arXiv:2308.07308.

\bibitem[{Radosevich and Halloran(2025)}]{radosevich2025mcpsafetyauditllms}
Radosevich, B.; and Halloran, J. 2025.
\newblock MCP Safety Audit: LLMs with the Model Context Protocol Allow Major
  Security Exploits.
\newblock arXiv:2504.03767.

\bibitem[{Subramani, Suresh, and Peters(2022)}]{subramani2022extracting}
Subramani, N.; Suresh, N.; and Peters, M.~E. 2022.
\newblock Extracting latent steering vectors from pretrained language models.
\newblock \emph{arXiv preprint arXiv:2205.05124}.

\bibitem[{Suo(2024)}]{suo2024signedpromptnewapproachprevent}
Suo, X. 2024.
\newblock Signed-Prompt: A New Approach to Prevent Prompt Injection Attacks
  Against LLM-Integrated Applications.
\newblock arXiv:2401.07612.

\bibitem[{Yang et~al.(2018)Yang, Qi, Zhang, Bengio, Cohen, Salakhutdinov, and
  Manning}]{yang-etal-2018-hotpotqa}
Yang, Z.; Qi, P.; Zhang, S.; Bengio, Y.; Cohen, W.; Salakhutdinov, R.; and
  Manning, C.~D. 2018.
\newblock {H}otpot{QA}: A Dataset for Diverse, Explainable Multi-hop Question
  Answering.
\newblock In Riloff, E.; Chiang, D.; Hockenmaier, J.; and Tsujii, J., eds.,
  \emph{Proceedings of the 2018 Conference on Empirical Methods in Natural
  Language Processing}, 2369--2380. Brussels, Belgium: Association for
  Computational Linguistics.

\bibitem[{Yao, Lou, and Qin(2024)}]{yao2024poisonprompt}
Yao, H.; Lou, J.; and Qin, Z. 2024.
\newblock Poisonprompt: Backdoor attack on prompt-based large language models.
\newblock In \emph{ICASSP 2024-2024 IEEE International Conference on Acoustics,
  Speech and Signal Processing (ICASSP)}, 7745--7749. IEEE.

\bibitem[{Yao et~al.(2025)Yao, Shi, Chen, Jiang, Wang, and
  Qin}]{yao2025controlnet}
Yao, H.; Shi, H.; Chen, Y.; Jiang, Y.; Wang, C.; and Qin, Z. 2025.
\newblock ControlNET: A firewall for rag-based LLM system.
\newblock \emph{arXiv preprint arXiv:2504.09593}.

\bibitem[{Zhang et~al.(2024)Zhang, Li, Du, Zhang, Zhao, Feng, and
  Yin}]{zhang2024hijackrag}
Zhang, Y.; Li, Q.; Du, T.; Zhang, X.; Zhao, X.; Feng, Z.; and Yin, J. 2024.
\newblock HijackRAG: Hijacking Attacks against Retrieval-Augmented Large
  Language Models.
\newblock \emph{arXiv preprint arXiv:2410.22832}.

\bibitem[{Zhong et~al.(2025)Zhong, Chen, Wang, McCall, Titzer, Miller, and
  Gibbons}]{zhong2025rtbas}
Zhong, P.~Y.; Chen, S.; Wang, R.; McCall, M.; Titzer, B.~L.; Miller, H.; and
  Gibbons, P.~B. 2025.
\newblock Rtbas: Defending llm agents against prompt injection and privacy
  leakage.
\newblock \emph{arXiv preprint arXiv:2502.08966}.

\bibitem[{Zou et~al.(2024{\natexlab{a}})Zou, Wang, Kolter, and
  Fredrikson}]{zou2024universal}
Zou, A.; Wang, Z.; Kolter, J.~Z.; and Fredrikson, M. 2024{\natexlab{a}}.
\newblock Universal and transferable adversarial attacks on aligned language
  models, 2023.
\newblock \emph{URL https://arxiv. org/abs/2307.15043}, 19.

\bibitem[{Zou et~al.(2024{\natexlab{b}})Zou, Geng, Wang, and
  Jia}]{zou2024poisonedrag}
Zou, W.; Geng, R.; Wang, B.; and Jia, J. 2024{\natexlab{b}}.
\newblock PoisonedRAG: Knowledge poisoning attacks to retrieval-augmented
  generation of large language models.
\newblock \emph{arXiv preprint arXiv:2402.07867}.

\end{thebibliography}

\input{ReproducibilityChecklist}

\end{document}